\documentclass[final]{cvpr}

\usepackage{times}
\usepackage{epsfig}
\usepackage{graphicx}
\usepackage{amsmath}
\usepackage{amssymb}

\usepackage[ruled]{algorithm2e} %
\usepackage{booktabs} %
\usepackage{graphicx}
\usepackage{makecell}
\usepackage{multirow}
\usepackage{wrapfig}
\usepackage{soul}

\usepackage{color}
\usepackage{steinmetz}
\usepackage{amssymb}
\usepackage{amsmath}
\usepackage{mathtools}
\usepackage{algorithmic}
\usepackage{newlfont} %
\usepackage{fixltx2e}
\usepackage[english]{babel} %
\usepackage{enumitem}
\usepackage{microtype}
\usepackage{floatflt}
\usepackage{booktabs} %
\usepackage{subcaption}
\usepackage{epsfig}
\usepackage[dvipsnames]{xcolor}
\usepackage{tabularx}
\usepackage{ragged2e}
\usepackage{arydshln}

\newcommand{\mypara}[1]{\vspace{-4mm}\paragraph{#1}}

\newcommand{\tpm}{$\pm$}

\usepackage[pagebackref=true,breaklinks=true,colorlinks,bookmarks=false,hypertexnames=false]{hyperref}

\def\cvprPaperID{7452} 
\def\confYear{CVPR 2021}

\begin{document}

\title{What Can Style Transfer and Paintings Do For Model Robustness?}


\author{ 
  {\normalsize
  Hubert Lin$^{1}$ \qquad
  Mitchell van Zuijlen$^{2}$ \qquad
  Sylvia C. Pont$^{2}$ \qquad
  Maarten  W.A. Wijntjes$^{2}$ \qquad
  Kavita Bala$^{1}$
  } \\
  {\small
  $^{1}$Cornell University \qquad
  $^{2}$Delft University of Technology
  }\\
  {\tt \small
  $^{1}$\{hubert, kb\}@cs.cornell.edu \qquad
  $^{2}$\{m.j.p.vanzuijlen, s.c.pont, m.w.a.wijntjes\}@tudelft.nl
  }
}

\maketitle

\begin{figure*}[t]
  \centering
  \includegraphics[width=\linewidth]{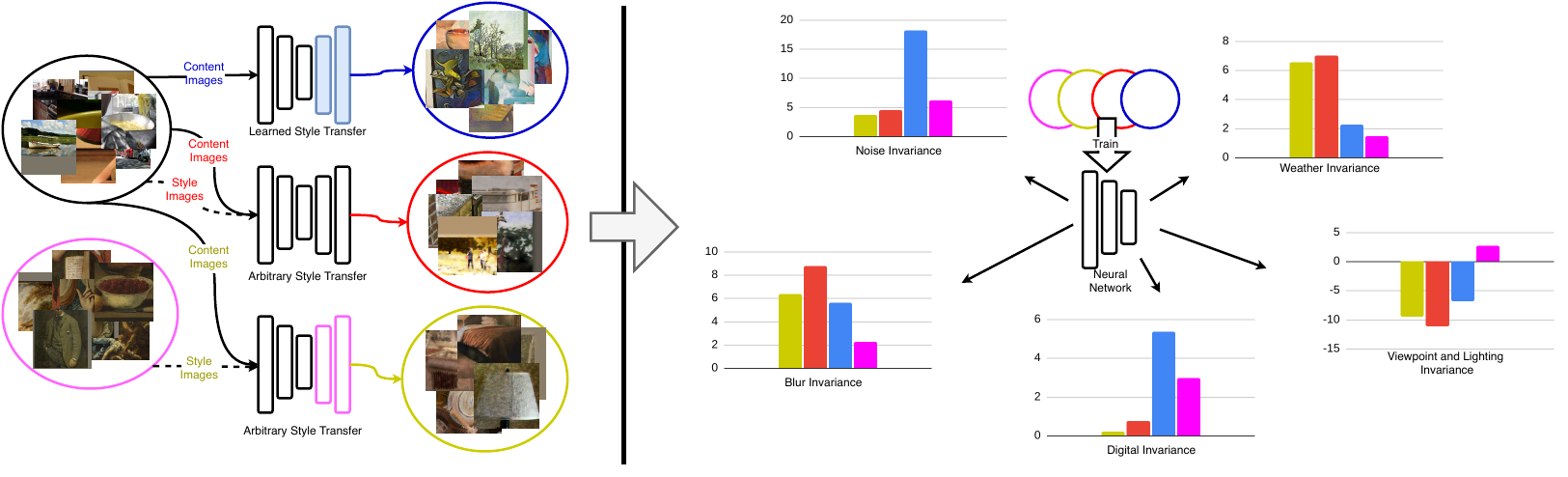}

  \caption{ \textbf{What invariances are learned from real and fake paintings?}
  Left: Natural photographs (\textcolor{black}{black}), paintings
  (\textcolor{Thistle}{magenta}), and stylized photographs
  (\textcolor{olive}{olive}/\textcolor{red}{red}/\textcolor{blue}{blue}) from
  the Materials dataset (Section \ref{sec:data}), Right: Relative robustness to
  various types of transformations for models trained with different sets of
  images with respect to a model trained on only natural photos. Stylization
  algorithms can transform photographs into painting-like images, but it is not
  clear that models will learn the same invariances from these images. This
  paper explores a series of hypotheses to understand the different ways in
  which style transfer and paintings improve model robustness. }

  \label{fig:teaser}

\end{figure*}

\begin{abstract}

A common strategy for improving model robustness is through data augmentations.
Data augmentations encourage models to learn desired invariances, such as
invariance to horizontal flipping or small changes in color. Recent work has
shown that arbitrary style transfer can be used as a form of data augmentation
to encourage invariance to textures by creating painting-like images from
photographs.  However, a stylized photograph is not quite the same as an
artist-created painting. Artists depict perceptually meaningful cues in
paintings so that humans can recognize salient components in scenes, an emphasis
which is not enforced in style transfer. Therefore, we study how style transfer
and paintings differ in their impact on model robustness. First, we investigate
the role of paintings as style images for stylization-based data augmentation.
We find that style transfer functions well even without paintings as style
images. Second, we show that learning from paintings as a form of perceptual
data augmentation can improve model robustness.  Finally, we investigate the
invariances learned from stylization and from paintings, and show that models
learn different invariances from these differing forms of data.  Our results
provide insights into how stylization improves model robustness, and provide
evidence that artist-created paintings can be a valuable source of data for
model robustness. Code and data are available at:
\small{\url{https://github.com/hubertsgithub/style\_painting\_robustness}}
\vspace{-2mm}

\end{abstract}

\section{Introduction}
\label{section:introduction}

  Model robustness can be defined as the capability of a model to generalize to
  unseen image distributions. These can be the  result of real-world effects, like
  weather and camera noise \cite{common_corr}, adversarial noise
  \cite{adv_noise}, or distribution shifts due to differences in environments in
  which the images are captured. The performance of standard recognition models
  can degrade drastically in these settings, but robust models are critical for
  applications such as self-driving or medical diagnostics.

  A common strategy is to improve generalization through data augmentation
  \cite{cutmix, cutout, augmix, adv_noise}. Conventional data augmentation
  applies transformations to encourage invariance to heuristic rules (e.g.,
  flipping for invariance to image mirroring). Recent work has found that image
  stylization can encourage models to learn invariance to texture \cite{SIN}.
  While style transfer has focused on visual fidelity \cite{nst}, we argue that
  current style transfer models do not yet fully capture the essence of artistic
  paintings. For example, a family of style transfer algorithms act by
  manipulating feature distributions to create a stylized photo which
  holistically mimics a painting \cite{demystifying} -- in effect, mid-level
  textures are manipulated in the stylized photo. However, paintings are
  more than a style filter applied to a photo.  An artist can
  choose lighting, contours, and scene context to convey realism in important
  scene regions while foregoing perceptual details less important areas. This
  artistic manipulation can affect our perceptual understanding of the scene.

  In this paper, we explore a series of hypotheses to understand how style
  transfer and paintings impact model robustness. Fig. \ref{fig:teaser}
  illustrates that various types of images can differently affect model
  robustness.  First, we examine how style images play a role in
  stylization-based data augmentation in Section \ref{sec:style}. Second, we
  investigate the role of paintings as a form of training data, and contrast it
  to other artforms such as sketches in Section \ref{sec:painting}. Finally, we
  probe models to empirically understand their learned invariances, and discuss
  how style transfer and artistic paintings can contribute to robust natural
  image recognition models in Section \ref{sec:painting_vs_style}. Our
  contributions are:

\begin{itemize}
    \vspace{-2.5mm}
  \item We demonstrate that arbitrary style transfer can be used as effective
    data augmentation even without painting style images. We attribute their
    effectiveness to the diversity of style images rather than artistic style.
    \vspace{-2mm}
  \item We argue that paintings can be considered a form of perceptual data
    augmentation, and demonstrate that it can improve model robustness. We
    contrast paintings with other forms of art such as sketches.
    \vspace{-2mm}
  \item We explore the invariances learned from arbitrary style transfer,
    learned artistic style transfer, and paintings.
    We find that models do not learn the same invariances from stylized photos
    and paintings, and show that the learned invariances are complementary.
\end{itemize}

\section{Related Work}
\label{sec:related}

\paragraph{Model Robustness.}

Recent work in robustness for CNNs has focused on both adversarial robustness
\cite{adv_rob} as well as robustness to real-world transformations
\cite{common_corr, dnn_gen}. This view of model robustness is human-centric,
where the settings considered are those where the human visual system has been
shown to be robust (e.g., \cite{fmd, dnn_gen, SIN}), rather than enforcing model
robustness under arbitrary settings. A related line of work is in domain
generalization, where the task is to generalize to unseen domains, (e.g.,
\cite{gulrajani2020search, mmld, li2019episodic, pacs}), by learning a shared
representation on a set of seen domains. While a common justification for domain
generalization is model robustness, domain generalization is subtly different.
Domain generalization algorithms assume the target domain is unspecified, and do
not rely on domain-specific signals at inference time.  However, robust natural
image recognition can benefit from learning from natural images directly.

\mypara{Data Augmentation.} Data augmentations are transformations applied to
images to enforce useful model invariances. Beyond basic transformations like
flipping, recent work in data augmentation has focused on more
complex augmentations such as image occlusion \cite{cutout}, class-mixing
\cite{cutmix}, and compositions of transformations \cite{augmix}.  Data-driven
augmentations such as adversarial or stylization transformations
\cite{adv_train, adv_noise, adain} can also be used to model nuanced
invariances.

\mypara{Style Transfer.} Style transfer aims to transform photos into
painting-like images by transferring artistic styles. While increasing attention
has been given to arbitrary style transfer (e.g., \cite{adain, etnet, avatarnet,
tpfr, distill_style}) which aims to efficiently transfer unseen styles, artist-specific style
transfer models (e.g., \cite{sacl, kotovenko2019content}) are typically able to better
capture nuances from a collections of images. Beyond its role as a tool for
artistic creation, stylization has also been used as a form of data
augmentation to enforce invariances to textures \cite{SIN}, as well as
regularization for tasks such as human re-identification
\cite{jin2020style}.

\section{Preliminaries}


\subsection{Evaluating Robustness}
\label{sec:eval}

We evaluate robustness to common image corruptions and distribution shifts from
the training distribution. These settings serve as a proxy for real-world
robustness.  Furthermore, the behavior of models on these scenarios gives us
insight into the invariances learned -- for example, a model which is robust to
noise has likely learned to be more invariant to (i.e., to rely little on)
high-frequency signals in an image.  All experiments use an ImageNet-pretrained
ResNet18 architecture, and results are averaged over three independent runs.
For complete training details and experiments with alternative architectures,
please refer to the supplementary.

\begin{figure}[!ht]
    \centering
    \includegraphics[width=\linewidth]{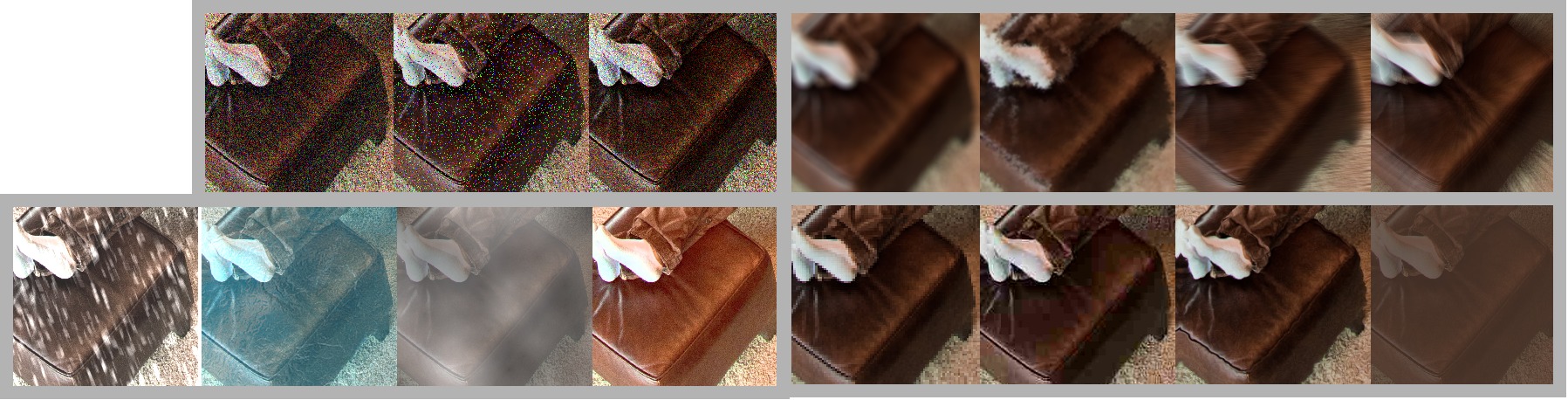}
    \\

    \caption{\textbf{Image Corruptions.} Top-Left to Bottom-Right:
    Noise($\times$3), Blur($\times$4), Weather($\times$4), Digital($\times$4).}
    \label{fig:common_corr}
\end{figure}

\mypara{Common Image Corruptions.} Common image corruptions are inspired by
transformations that can be encountered in real-world settings
\cite{common_corr}.  There are 15 corruptions which span 4 broad categories
(noise, blur, weather, and digital) with 5 severity levels per
corruption. We use the released code to corrupt our test images.  Figure
\ref{fig:common_corr} illustrates these corruptions. For each corruption, we
compute the mean accuracy over each severity, and then compute the mean over
each set of broad corruption categories $C$.  Given a model $\Theta$, the mean
corruption accuracy is:

{\small

\begin{align}
  &\text{Acc}_\text{Mean}(\Theta) = \dfrac{1}{4}\sum\limits_{C}
  \text{Acc}_C(\Theta) \\
  \nonumber &\text{where } \text{Acc}_C(\Theta) = \dfrac{1}{5n_C}\sum\limits_{corr \in C}
  \sum\limits_{s=1}^{5} \text{Acc}(\Theta,  \mathcal{D}_{corr,s}) 
\end{align}

}%

$\mathcal{D}_{corr, s}$ denotes the test dataset of images transformed by
corruption $corr$ with severity $s$.

\mypara{Small Distribution Shifts.} Out-of-distribution photographs will be used
to evaluate robustness to small domain shifts not unlike the domain shifts that
models must overcome when they are used in different real world environments.
For the PACS dataset, we use a subset of the YFCC100M dataset \cite{yfcc100m} as
the out-of-distribution test set. This subset is curated by downloading 100
images per class and then manually filtering to remove irrelevant retrievals
down to 50 images per class. This test set is released for reproducibility.  For
the Materials dataset, we use the Flickr Material Database (FMD) \cite{fmd} as
the out-of-distribution test set.

\subsection{Datasets}
\label{sec:data}

We select datasets which contain both photographs and paintings, and conduct
experiments across two recognition tasks (object classification and material
classification). 

\mypara{Object Classification.}
We use the PACS dataset \cite{pacs} which consists
of 10K images across 7 categories and 4 domains (photographs,
paintings, cartoons, and sketches). 

\mypara{Material Classification.}

We construct a dataset from existing large-scale photograph datasets
\cite{opensurfaces, minc, cocostuff}, and a large-scale painting dataset with
material annotations \cite{van2020database}. We will refer to this dataset as
`Materials'. This dataset consists of 120K images across 10 categories and 2
domains (photographs and paintings).  See supplementary for
details. \cite{SIN} found that stylization-based augmentation can reduce bias
towards textures, but material recognition relies on texture understanding
\cite{adelson2001seeing}. As such, it is interesting to explore whether
stylization can improve robustness for this task.

\subsection{Notation}

Some common notation used throughout is given here. Let $\mathcal{D}_{n}$ be a
set of natural photographs and $\mathcal{D}_{p}$ be a set of paintings. For each
image $x$, its class label is denoted by $y_x$. Finally, let $l(\hat{y}, y)$ denote the cross entropy loss.

\section{Style Transfer as Data Augmentation}
\label{sec:style}

Style transfer aims to transform the style of an image into the style of another
set of images \cite{nst}. There is evidence \cite{SIN} that
training on stylized images \cite{adain} can improve object recognition on ImageNet by
encouraging networks to focus more on shape than texture. In this
view, we can consider style transfer as a form of data augmentation. Style
transfer is often applied with painting style images from datasets such as
Wikiart \cite{artgan, wikiart}. In its role as a tool to mimic artistic creation, this
is certainly appropriate. However, in its role as a form of data augmentation,
it is not strictly necessary for the style images to be paintings. Indeed,
arbitrary stylization methods can be applied to any pair of content and style
images (hence `arbitrary'). Although work such as \cite{SIN} utilize style
transfer in the conventional manner with painting styles, it's important to ask
whether models can learn robust invariances from \emph{photo} style images alone.


To answer this question in a general way, we experiment with three
representative deep-learning based arbitrary style transfer methods. Each of
these methods act in deep feature space, but follow a different paradigm: AdaIN
\cite{adain} transfers style by matching the mean and standard deviation of
features, ETNet \cite{etnet} iteratively refines a stylized image by computing
residual error maps, and TPFR \cite{tpfr} transfers style by recombining
features in the content image to match those of the style image. We explore the
following:

\begin{itemize}
    \vspace{-2mm}
  \item \textbf{Hypothesis H1.} Painting styles are necessary for
    stylization-based augmentation to improve robustness.
    \vspace{-2mm}
    \item \textbf{Hypothesis H2.} Style image diversity is important.
\end{itemize}

\subsection{Are Painting Style Images Necessary?}

We experiment with: (a) a network trained with photos
plus photos stylized by paintings and (b) a network trained with photos plus
photos stylized by other photos. We will refer to (b) as ``intradomain
stylization'' as photos are being stylized by other photos from within the same
domain. For reference, we also consider (c) a network trained with photos alone
(no stylization). Specifically, let $\phi(x, x_s)$ be an arbitrary stylization algorithm
which stylizes content image $x$ with style image $x_s$.  For a network
$\Theta$, the objectives are given by:

{\small

\begin{align}
  \nonumber&\text{(a)}\min_\Theta \mathbb{E}_{x, x_s \sim
  \mathcal{D}_{n},\mathcal{D}_{\textcolor{red}p}}\big\lbrack
  \dfrac{1}{2}\big(l(\Theta(x), y_x) + l(\Theta(\phi(x, x_s)), y_x)\big)\big\rbrack \\
  \nonumber&\text{(b)}\min_\Theta \mathbb{E}_{x, x_s \sim
  \mathcal{D}_{n},\mathcal{D}_{\textcolor{red}n}}\big\lbrack
  \dfrac{1}{2}\big(l(\Theta(x), y_x) + l(\Theta(\phi(x, x_s)),
  y_x)\big)\big\rbrack \\
  &\text{(c)}\min_\Theta \mathbb{E}_{x\sim
  \mathcal{D}_{n}}\big\lbrack l(\Theta(x), y_x)\big\rbrack
\end{align}

}%


In practice, we approximate the objectives by sampling $x_s$ once for each $x$
instead of minimizing over all independent combinations of $x$ and $x_s$.

\begin{figure}[!ht]
    \centering
    \includegraphics[width=\linewidth]{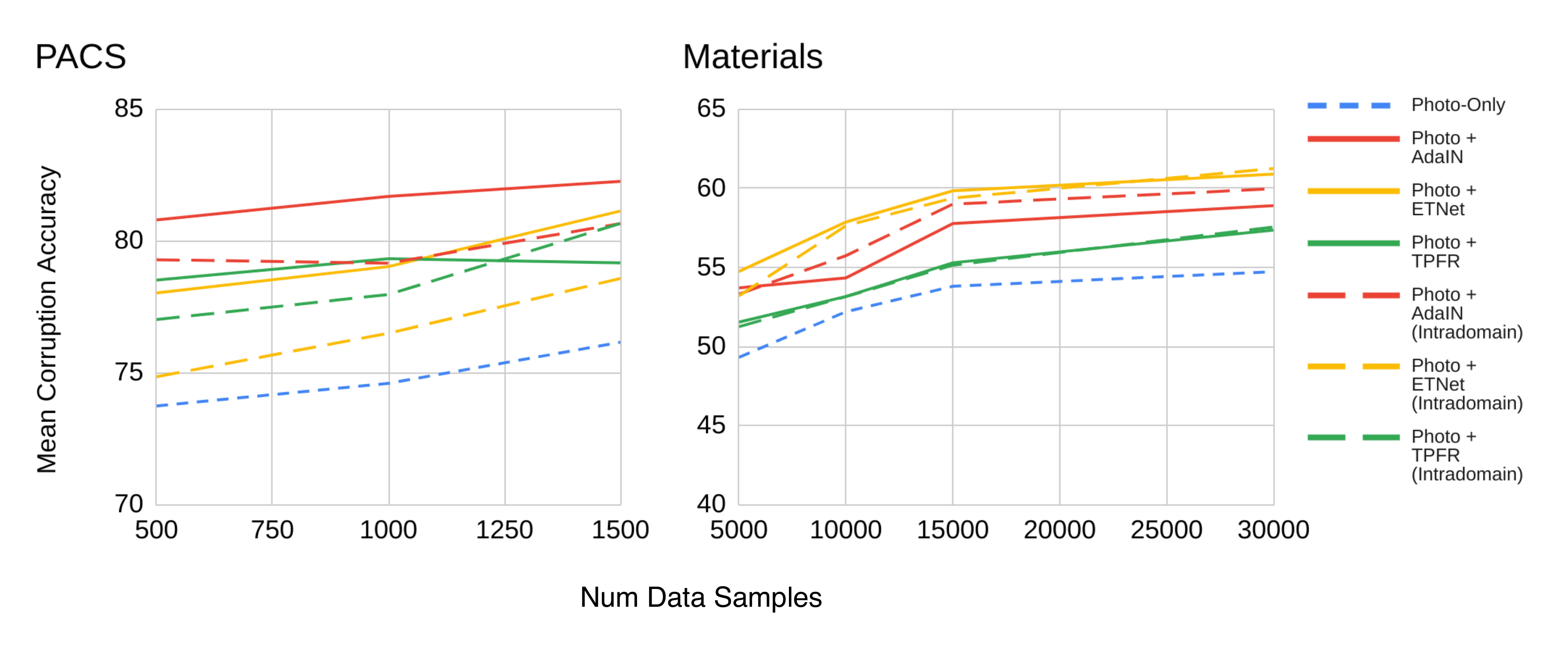}

    \caption{ \textbf{Stylization: Painting vs Photo Styles.} Left: PACS, Right:
    Materials.   In general, intradomain stylization
    (\textcolor{red}{red}/\textcolor{ForestGreen}{green}/\textcolor{brown}{yellow})
    improves robustness over no stylization (\textcolor{blue}{blue}).  Further,
    when sufficient data is available (Materials), intradomain stylization
    (dashed lines) results in similar robustness gains to conventional painting
    stylization (solid lines). This means that paintings are not uniquely
    responsible for robustness gains from stylization.}

    \label{fig:intradomain_style}
\end{figure}

The results are shown in Fig. \ref{fig:intradomain_style}.  Across both PACS and
Materials, we find that intradomain stylization significantly improves
robustness over the photo-only baseline. With a large dataset (Materials), we
find that intradomain stylization can meet or even exceed the performance of
conventional painting-based stylization. Thus, in contrast to common practice,
stylization-based data augmentation does \ul{not} need painting style images.
This finding is also supported by recent work which shows that online feature
moment matching across different training images is an effective form of data
augmentation \cite{moex} (which we can frame as roughly equivalent to
intradomain stylization with AdaIN), and work which shows stylization with
images from non-painting domains (including intradomain stylization) can be
useful for domain generalization \cite{frust_dg}. We have shown
explicitly here that intradomain stylization can replace painting stylization
for robust natural image recognition when enough data is available.


\noindent \textbf{Answer to H1:}  
\emph{Intradomain stylization can improve network robustness to an extent that
is comparable to painting stylization \ul{when there is sufficient data} -- that
is, paintings do not play a unique role when arbitrary style transfer is used as
data augmentation.}

\subsection{The Role of Style Diversity \label{sec:intraclass}}

The finding that intradomain stylization can be comparable to painting
stylization leads to the hypothesis that it is the diversity in image statistics
between style and content images that plays a key role.
For example, consider AdaIN -- the extent to which images are transformed by
stylization depends on the magnitude of the difference in feature distribution
moments between the content image and the style image.  This is why intradomain
stylization is comparable to painting stylization on a large dataset like
Materials.

We test this hypothesis by restricting the style photo for intradomain
stylization to be drawn from images that share the same class label as the
content image. With this restriction, the style images are likely to be more
similar to the content image given that they share similar semantic content. Let
$\mathcal{D}_n^{y}$ be the subset of natural photographs with class label $y$.
Then, the objective is given by:




{\small

\begin{align}
  &\min_\Theta  \mathbb{E}_{x\sim
  \mathcal{D}_{n}}\bigg\lbrack \mathbb{E}_{x_s \sim
  \mathcal{D}_{n}^{\textcolor{red}{y_x}}} \big\lbrack
  \dfrac{1}{2}\big(l(\Theta(x), y_x) + l(\Theta(\phi(x, x_s)),
  y_x)\big)\big\rbrack \bigg\rbrack
\end{align}

}%

\begin{figure}[!ht]
    \centering
    \includegraphics[width=\linewidth]{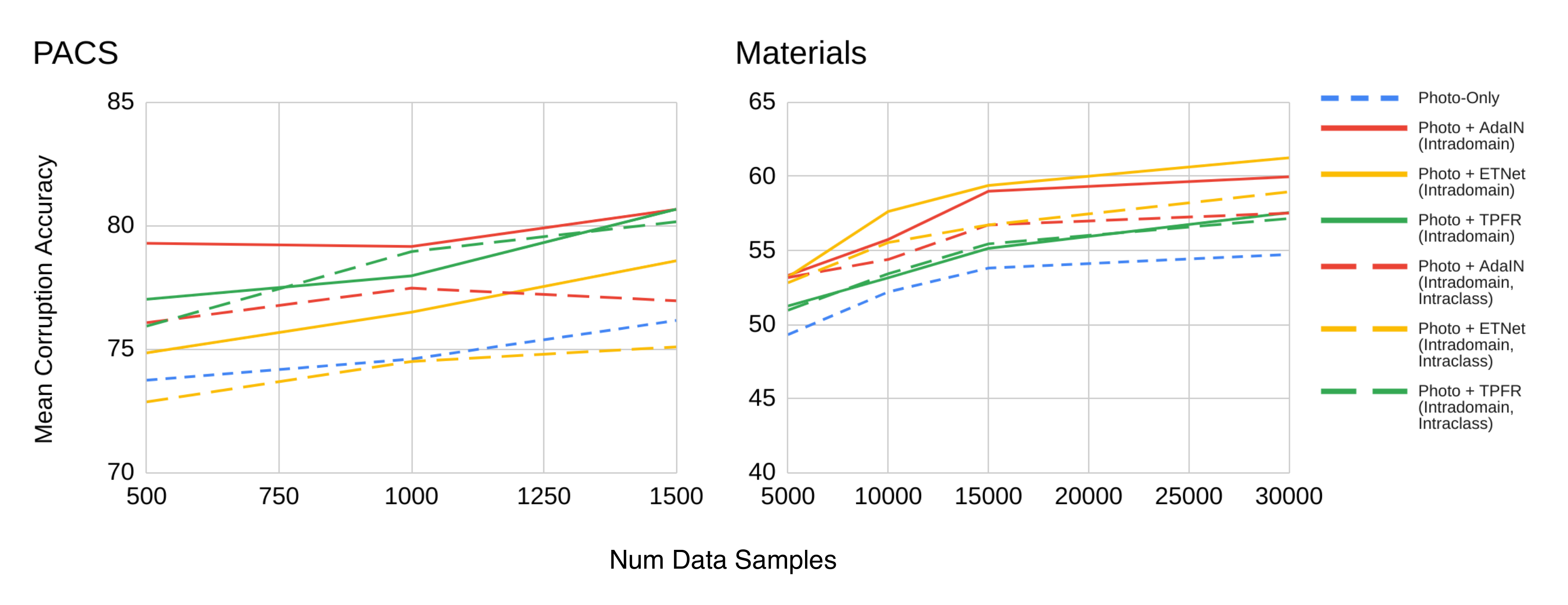}

    \caption{ \textbf{Stylization: Unrestricted vs Intraclass Styles.}
    Left: PACS, Right: Materials.  Across both datasets, restricting
    style images to the class as content images (dashed lines) results in
    smaller robustness gains compared to unrestricted stylization
    (solid lines). This reduction in robustness is explained by the reduction in
    diversity between content images and style images.   
    }

    \label{fig:intraclass_style}
\end{figure}

In general, we find that this restriction does indeed reduce the effectiveness
of intradomain stylization (Fig. \ref{fig:intraclass_style}). As an exception,
TPFR does not appear to rely heavily on the choice of style images.  This can be
explained by the adversarial loss used in TPFR -- the decoder is trained
explicitly to fool a style discriminator that discriminates between stylized
images and real paintings during training.  Therefore, it is possible that the
decoder is encoding painting-like style signals regardless of the style image
used. This also suggests that a style transfer algorithm which explicitly
transfers painting styles can be useful instead of relying on a diverse style
dataset during training (we explore this in Section
\ref{sec:painting_vs_style}). In general, biases in stylization models can
contribute to improved robustness independently of style images.

\noindent \textbf{Answer to H2:} \emph{Access to style images which are diverse
\ul{with respect to content images} is key for stylization-based augmentation.}
Against conventional wisdom, style images need not contain statistics that
manifest as visible textures or artistic style per se.  As long as each style
image is sufficiently different from its corresponding content image, it will
suffice.  ``Sufficiently different'' means ``depicting different semantic
content'' in our analysis here. Interestingly, we found that style differences
measured by the Gram matrix distance between a stylized image and its original
counterpart do not correlate with robustness (see supplementary) -- further
analysis is left for future work.


\section{Paintings as Perceptual Data Augmentation}
\label{sec:painting}

In Section \ref{sec:style}, we found that stylization as data augmentation works
well as long as the set of style images are diverse. This diversity does not
necessarily depend on the image statistics found specifically in paintings.
If sufficiently diverse mid-level statistics is found by stylization with
photos, then perhaps photos can fulfill the role of paintings entirely.

Instead, we argue that paintings are more than just a set of mid-level style
features overlaid on top of a photograph.  
Our key insight is that perceptually realistic paintings can be considered a
form of `perceptual' data augmentation. Unconstrained by physical reality,
artists are free to depict varying level of perceptual realism \cite{cavanagh}.
Paintings are \emph{perceptually} realistic in regions where the artist has
deemed viewer attention should be focused. For example, a painting of a giraffe
might include perceptually relevant details on the giraffe itself while the
background is depicted in an less realistc and more abstract manner. In a
collection of paintings, important cues for objects or materials of interest are
depicted frequently in a perceptually sound manner while unimportant details are
abstracted away.

Even so, the domain shift between paintings and photos can be problematic, and
it is likely that models trained on paintings will fail to perform well on photos if
domain shift is not accounted for. Furthermore, many of the arguments made for
paintings above can also apply to other artforms, and it is interesting to
consider alternatives. We explore the following:

\begin{itemize}
    \vspace{-2mm}
    \item \textbf{Hypothesis H3.} (a) Learning from paintings improves natural
      image robustness after accounting for domain shift, and (b) this
      improvement is greater than that found from photos alone. 
      \vspace{-2mm}
    \item \textbf{Hypothesis H4.} Other artforms can encode similar invariances
      to paintings.
\end{itemize}



\subsection{Learning Robust Natural Image Recognition From Paintings \label{sec:domain_fc}}

A classifier trained directly on both photos and paintings is required to learn
boundaries that satisfy both of these domains. Consequently, the accuracy on
photographs can suffer. Since our goal is to train a robust model for
natural image classification, we alleviate this by considering two alternatives:
(a) a shared feature extractor with multiple domain-specific classifiers
(multitask learning) or (b) a photo-only classifier that is finetuned after
shared feature learning. For reference, we also consider the default option of
training (c) a joint classifier on both photos and paintings.  Specifically, let
$\Theta_{f}$ be a feature extractor (i.e., ResNet18 without the final fully
connected layer).  Let $\eta$ be a linear classifier (i.e., a fully connected
layer). Then the objective for (a) is given by: 

\begin{align}
  \nonumber&\min\limits_{\Theta_f, \eta_n, \eta_p} \mathbb{E}_{x_n, x_p \sim
  \mathcal{D}_{n},\mathcal{D}_{p}}\big\lbrack
  \dfrac{1}{2}\big(l((\eta_{\textcolor{red}n} \circ
  \Theta_f)(x_{\textcolor{red}n}), y_{x_n}) + \\ &
  \hspace{3.9cm} l((\eta_{\textcolor{red}p} \circ \Theta_f)(x_{\textcolor{red}p}),y_{x_p})\big)\big\rbrack
\end{align}

\noindent For (b), two objectives are optimized sequentially:

\begin{align}
  \text{(i) }\nonumber&\min\limits_{\Theta_f, \eta_n} \mathbb{E}_{x_n, x_p \sim
  \mathcal{D}_{n},\mathcal{D}_{p}}\big\lbrack
  \dfrac{1}{2}\big(l((\eta_{\textcolor{red}n} \circ
  \Theta_f)(x_{\textcolor{red}n}), y_{x_n}) + \\ &
  \nonumber\hspace{3.5cm} l((\eta_{\textcolor{red}n} \circ
  \Theta_f)(x_{\textcolor{red}p}),y_{x_p})\big)\big\rbrack\\
  \text{(ii) }&\min\limits_{\eta_n}
  \mathbb{E}_{x_n\sim\mathcal{D}_{n}}\big\lbrack l((\eta_{\textcolor{red}n} \circ
  \Theta_f)(x_{\textcolor{red}n}), y_{x_n}) \big\rbrack
  \label{eq:joint_ft}
\end{align}

\noindent For (c), the objective is simply Eq. \ref{eq:joint_ft}(i). In all
cases, the model defined by $(\eta_n \circ \Theta_f)$ is used at inference time.
Both options (a) and (b) allow paintings to be used for feature learning while
keeping the inference classifier specific to photos.

\begin{figure}[!ht]
    \centering
    \includegraphics[width=\linewidth]{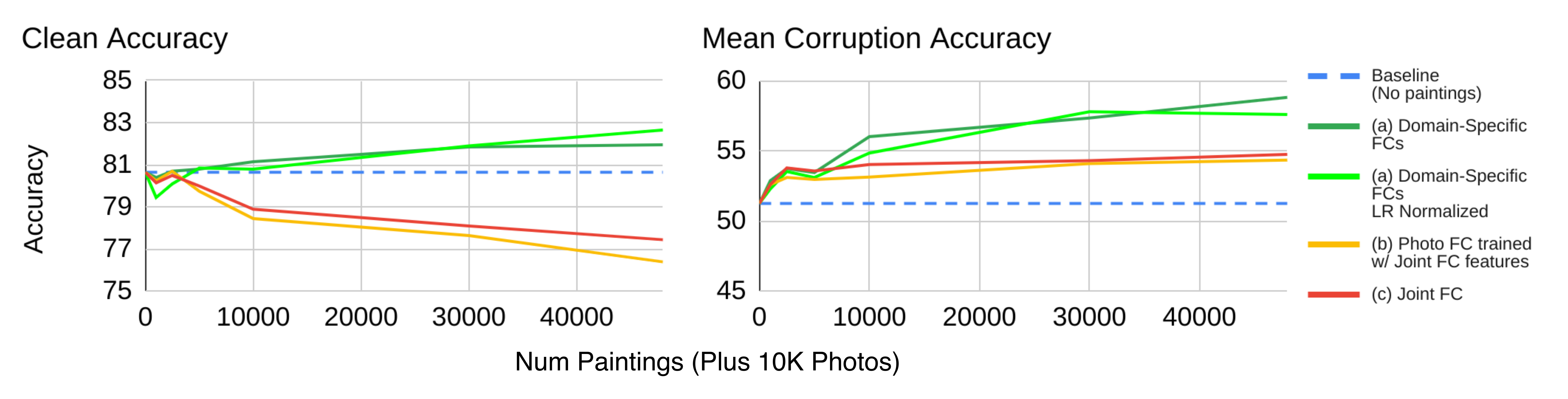}

    \caption{\textbf{Learning from Paintings.} Left: Clean Accuracy, Right:
    Corruption Accuracy. Domain-specific classifiers
    (\textcolor{ForestGreen}{green}) result in the highest robustness while also
    improving clean accuracy. ``LR normalized'' refers to fixed effective
    learning rates to account for additional gradients from the extra classifier
    head. Even without accounting for domain shifts, training with paintings
    improves robustness (\textcolor{red}{red}/\textcolor{brown}{yellow}).
    Results are on Materials. 
    }


    \label{fig:domain_specific}
\end{figure}

Results are summarized in Fig. \ref{fig:domain_specific}.  Despite domain
differences between photos and paintings, the default classifier (c) has
improved robustness over a classifier that is trained on photos alone. A
finetuned classifier (b) does not yield much improvement over the default option
(c), while domain-specific classifiers (a) do yield significant improvement. This
suggests that paintings are useful for feature learning since they can guide the
feature extractor towards perceptually relevant features, but constraining the
feature space to jointly separate photos and paintings across different classes
can restrict the breadth of learned features. The clean accuracy of a joint
classifier (finetuned or not) suffers since it can no longer rely on some
photo-specific features for classification.  We will use domain-specific
classifiers in remaining experiments unless otherwise specified.\footnote{We
experimented with domain-specific classifiers in the context of stylization, but
found they did not improve robustness over a joint classifier.}

\noindent \textbf{Answer to H3a:} \emph{Surprisingly, we find that
paintings can improve model robustness out-of-the-box \ul{without} accounting
for domain shift. However, accounting for domain shift with domain-specific classifiers
increases both clean accuracy and robustness significantly.}


\begin{figure}[!ht]
    \centering
    \includegraphics[width=\linewidth]{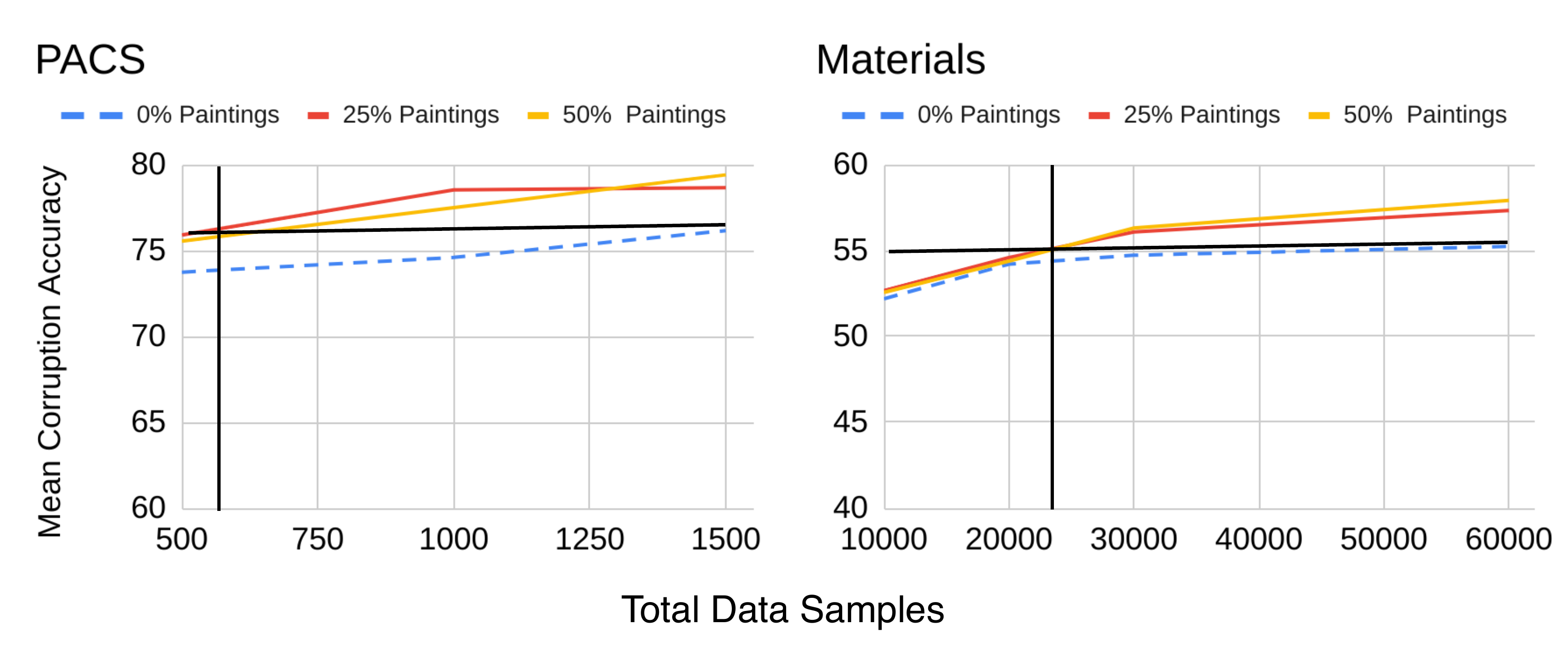}

    \caption{ \textbf{Trade-off Between Photos and Paintings.} Left: PACS,
    Right: Materials. For a fixed annotation budget, learning from both photos and
    paintings (\textcolor{red}{25\%}/\textcolor{brown}{50\%} paintings) results in higher robustness than photos
    alone (\textcolor{blue}{0\%} paintings), with $<$50\% of the total number of
    data samples annotated achieving the maximal robustness achieved
    by only photos. }

    \label{fig:materials_fixed_budget}
\end{figure}

To control for robustness gains from photos, we assume a 1:1 cost for
photos:paintings with a fixed annotation budget. Fig.
\ref{fig:materials_fixed_budget} shows that it is beneficial to allocate up to
50\% of any annotation budget for paintings with respect to model robustness. 

\noindent \textbf{Answer to H3b:} \emph{Using paintings is cost-effective -- annotating a combination of photos
and paintings results in higher robustness over photos alone for any fixed budget.}

\subsection{Paintings vs. Other Visual Artforms}

Many artforms are created with an artistic emphasis on perceptually important
cues. For example, a line sketch is an abstraction which focuses  on salient
contours to depict recognizable objects.  
While sketches are quite good at abstracting away unimportant signals, they also
abstracts away many realistic cues in favor of a sparse line-based
representation. In the following experiment, we consider models trained on
photographs with different visual artforms. 

Table \ref{table:robust} summarizes results across four datasets. We find that
robustness can be harmed by sparse visual representations like PACS line
sketches or DomainNet quickdraw. However, DomainNet sketches, which include more
realistic shading and detail, do improve robustness. This is aligned with our
expectation that the inclusion of perceptually relevant cues is important for
feature learning. VisDA renderings are untextured and shaded with a single
directional light source and ambient lighting. Similar to line sketches, we find
that these minimal renderings reduce model robustness.  

\noindent \textbf{Answer to H4:} \emph{Our results position paintings as a
unique artform for improving model robustness due to their fine balance between
perceptual realism and abstraction.}






\begin{table}[h!]
  \small
  \centering
  \begin{tabular}{l|l}
    \hline
    Training Data (\# Samples)     & \makecell[c]{Mean Corruption Acc (\%)}  \\
    \hline
    \multicolumn{2}{c}{\emph{Materials}}\\
    \hline
    Photo (30K) & 54.73\tpm0.25 \\
    Photo + \textbf{Painting} (15K + 15K) & \textbf{56.31}\tpm0.27 $(+)$ \\
    \hline
    \multicolumn{2}{c}{\emph{PACS}}\\
    \hline
    Photo (1500) & 76.16\tpm0.34 \\
    Photo + \textbf{Painting} (750 + 750) & \textbf{79.41}\tpm0.55 $(+)$\\
    Photo + Cartoon (750 + 750) & 75.38\tpm0.36 $(-)$ \\
    Photo + Sketch (750 + 750) & 73.85\tpm0.39 $(-)$ \\
    \hline
    \multicolumn{2}{c}{\emph{DomainNet \cite{domainnet}}}\\
    \hline
    Photo (120K) & 36.59\tpm0.12 \\
    Photo + \textbf{Painting} (90K + 30K) & \textbf{39.00}\tpm0.14 $(+)$\\
    Photo + Sketch (90K + 30K) & 37.57\tpm0.22 $(+)$ \\
    Photo + Clipart (90K + 30K) & 37.00\tpm0.07 $(+)$\\
    Photo + Quickdraw (90K + 30K) & 35.87\tpm0.20 $(-)$ \\
    Photo + Infograph (90K + 30K) & 34.60\tpm0.18 $(-)$ \\
    \hline
    \multicolumn{2}{c}{\emph{VisDA \cite{visda}}}\\
    \hline
    Photo (30K) & \textbf{65.97}\tpm0.33 \\
    Photo + Rendering (15K + 15K) & 63.90\tpm0.21 $(-)$\\
    \hline
  \end{tabular}

  \caption{ \textbf{Robustness from Different Artforms.} Paintings improve model
  robustness while more abstract artforms can reduce robustness. $(+)$/$(-)$~indicate whether
  an artform improves/reduces model robustness. \tpm~indicates standard
  deviation over 3 runs. }

  \label{table:robust}
\end{table}

\section{Do Stylized Images and Paintings Induce Similar Invariances?}
\label{sec:painting_vs_style}


As shown in Sections \ref{sec:style} and \ref{sec:painting}, both stylized
images and paintings can improve model robustness. We argued that paintings are
a form of perceptual data augmentation in which artists manipulate perceptual
cues to emphasize salient regions of scenes. However, it remains unclear whether
models are indeed learning perceptual invariances from paintings -- it is
possible that the robustness gains from paintings arise purely through their
mid-level image statistics and textures instead. If paintings are
improving robustness through different mechanisms than stylized photos, we can
expect different behavior from models trained on stylized photos and paintings.
To investigate how stylized photos and paintings act on model robustness, we
empirically probe models to understand their learned invariances. We explore the
following:

\begin{itemize}
    \vspace{-2mm}
    \item \textbf{Hypothesis H5.} Models trained on stylized photos and paintings
      learn different invariances to (a) common image corruptions and (b)
      viewpoint and lighting shifts, and so (c) models can learn complementary invariances by training on both paintings
      and stylized photos. 
    \vspace{-2mm}
    \item \textbf{Hypothesis H6.} Stylization injects high-frequency
      signals that improve model robustness.
\end{itemize}


\subsection{Probing Learned Invariances \label{sec:per_corr}}

To explore the relative invariances learned by different models, we consider the
behavior of models on various types of common image corruptions.  We also
consider behavior on out of distribution images -- in general, these images have
a different distribution of viewing angles, viewing scales, and lighting than
the original training photos.  We experiment with models trained on paintings
and AdaIN-stylized photos.  In addition to arbitrary style transfer, it is
natural to consider learned artistic style transfer.  We experiment with SACL
\cite{sacl}, which transfers the style of various artists independently with
separately trained models. We stylize each photo with a random artist to
parallel the real painting datasets which include multiple artists and styles. 




Behavior with respect to common corruptions is summarized in Table
\ref{table:per_corr}. Stylization and paintings both consistently improve
robustness to each form of common corruption. On average, SACL outperforms both
AdaIN and paintings, giving credence to an argument that stylization methods
with strong biases (i.e., learned styles) may be more practical than real
paintings or arbitrary stylization methods that depend on a diverse style set
(c.f. Section \ref{sec:intraclass}). Observe that the relative performance of
paintings fluctuates between datasets -- paintings outperform AdaIN on
noise and digital on Materials but underperform AdaIN on PACS. As discussed
earlier, a collection of paintings encodes perceptual invariances.  Since these
invariances are not agreed upon a priori for every painting, it follows that
\emph{a large set of paintings is required to adequately capture implicitly
encoded perceptual invariances}.  Finally, all methods are similarly invariant
to weather and digital transformations. This can be explained by their mid-level
statistics.  Weather transformations such as snow, fog, and frost are
effectively overlaid textures on an image while digital transformations such as
pixelate and elastic transform resemble the fuzzy boundaries found in both types
of images.  

\noindent \textbf{Answer to H5a:} \emph{Both stylization and paintings
improve robustness to various image corruptions. However, learned stylization
strictly outperforms paintings, suggesting that \ul{invariances from learned style
transfer supersedes those from paintings with respect to common corruptions}.}

\begin{table}[h!]
  \scriptsize
  \centering
  \begin{tabular}{l||c|c|c|c}
    \hline
      \!\!Method\!\! & \makecell[c]{Noise} & \makecell[c]{Blur} &
      \makecell[c]{Weather} &\makecell[c]{Digital}    \\
    \hline
    \multicolumn{5}{c}{\emph{Materials} (30K Samples/Domain)}\\
    \hline
    \!\!Photo-Only\!\! & 43.70\tpm0.65 & 58.76\tpm0.14 & 55.25\tpm0.33 &
    61.20\tpm0.69 \\
    \!\!Photo + AdaIN\!\! & \textcolor{red}{47.33}\tpm0.22 & \textbf{65.09}\tpm0.21 &
    \textbf{61.78}\tpm0.18 & \textcolor{red}{61.41}\tpm0.16\\
    \!\!Photo + SACL\!\! & \textcolor{blue}{\bf 61.87}\tpm0.16 & 64.36\tpm0.20 &
    57.49\tpm0.24 & \textbf{66.55}\tpm0.17 \\
    \!\!Photo + Painting\!\! & \textcolor{red}{49.82}\tpm0.56 & 61.03\tpm0.13 &
    56.69\tpm0.10 &
    \textcolor{red}{64.15}\tpm0.14 \\
    \hline
    \multicolumn{5}{c}{\emph{PACS} (1.5K Samples/Domain)}\\
    \hline
    \!\!Photo-Only\!\! & 62.64\tpm1.48 & 72.75\tpm0.04 & 83.24\tpm0.22 &
    86.33\tpm0.14 \\
    \!\!Photo + AdaIN\!\! & \textcolor{red}{70.17}\tpm1.70 & 81.18\tpm0.20 &
    88.37\tpm0.23 &
    \textcolor{red}{\bf 89.32}\tpm0.19 \\
    \!\!Photo + SACL\!\! & \textcolor{blue}{\bf 85.98}\tpm0.56 & \textbf{84.61}\tpm0.15 &
    \textbf{89.73}\tpm0.33 & 88.74\tpm0.48 \\
    \!\!Photo + Painting\!\! & \textcolor{red}{68.83}\tpm0.83 & 75.80\tpm0.95 &
    86.88\tpm0.66 & \textcolor{red}{87.07}\tpm0.14 \\
    \hline

  \end{tabular} 

  \caption{ \textbf{Per-Corruption Accuracy.} (\textcolor{blue}{blue}) SACL
  generally outperforms both AdaIN and paintings, particularly on noise.
  (\textcolor{red}{red}) Paintings can outperform AdaIN on some corruptions with
  a large dataset (Materials), but underperform when fewer images are available
  (PACS).  See main text for discussion. \tpm~indicates standard deviation over
  3 runs.} 

  \label{table:per_corr}
\end{table}

\begin{figure}[!ht]
    \centering
    \includegraphics[width=\linewidth]{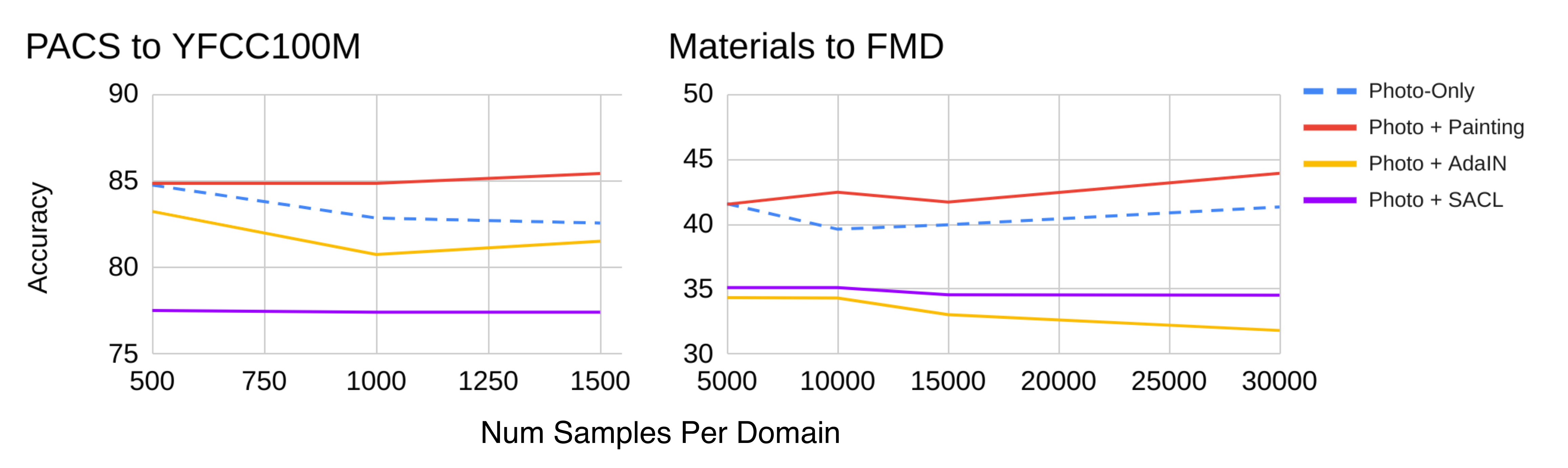}

    \caption{ \textbf{Out-of-Distribution Accuracy.} Left: PACS, Right:
    Materials. Training with paintings
    (\textcolor{red}{red}) improves robustness to out-of-distribution photos
    while training with stylized photos
    (\textcolor{RoyalPurple}{purple}/\textcolor{brown}{yellow}) hurts
    robustness. Paintings can improve invariance to viewpoints and lighting by
    encouraging models to focus on objects / materials of interest over
    background context. Stylization encourages overfitting, an effect which can
    be exacerbated with more training samples.}

    \label{fig:ood}
\end{figure}

Performance with respect to out-of-distribution images is summarized in Fig.
\ref{fig:ood}.  In striking contrast to the robustness against image corruption
results above, stylization consistently \emph{harms} robustness. The reduced
performance of stylization can be explained by model overfitting to view- or
lighting-specific signals in the original photo dataset, as the signals in
common between a clean photo and its stylized counterpart are seen twice as
often by the network during training. On the other hand, paintings are not
simply a transformed photograph, and thus do not suffer from this problem.  A
straightforward explanation of the robustness found through paintings is in the
differences in viewpoints and lighting depicted compared to photos due to
circumstance (that is, the paintings simply depict more diverse scenes than the
photos). However, paintings are constrained by cultural norms and artistic
conventions \cite{summers1981conventions, mamassian}, so it is unlikely that
artistic paintings contain a more diverse set of viewpoints than in-the-wild
photos.  Instead, \emph{we argue it is the emphasis on depicting regions of
interest with recognizable characteristics while de-emphasizing details in the
background that is helping networks to learn better viewpoint invariance from
paintings.}  The model is better able to learn to focus on the objects or
materials themselves over background context.  

\noindent\textbf{Answer to H5b:}  \emph{For viewpoint and lighting transformations found in
out-of-distribution images,  using \ul{stylization consistently hurts
performance} while using \ul{paintings consistently improves performance}.} 





Since the behavior of models trained on stylized photos and paintings are indeed
different, we explore whether models trained on both sources of data learn
complementary invariances, or if the differences result in conflicting behavior.
Our results in Table \ref{table:comp} suggests the former.

\noindent \textbf{Answer to H5c:} \emph{Training with both paintings and
stylized photos improves robustness in a complementary manner.}

\begin{table}[h!]
  \footnotesize
  \centering
  \begin{tabular}{l||c||c|c}
    \hline
      \!\!Method\!\! & \makecell[c]{MEAN} & \makecell[c]{Corr.} & \makecell[c]{OOD} \\
    \hline
    \multicolumn{4}{c}{\emph{Materials} (30K Samples/Domain)}\\
    \hline
    \!\!Photo-Only\!\! & 48.03\tpm0.21 & 54.73\tpm0.25 & 41.33\tpm0.62 \\
    \!\!Photo + SACL\!\! & 48.56\tpm0.45 & \textbf{62.67}\tpm0.03 & 34.54\tpm0.91 \\ 
    \!\!Photo + Painting\!\! & 50.92\tpm0.22 & 57.92\tpm0.09 &  \textbf{43.92}\tpm0.47 \\
    \!\!Photo + SACL + Painting\!\! & \textbf{51.49}\tpm0.69 & 61.47\tpm0.50 &
    41.50\tpm1.38 \\ 
    \hline
    \multicolumn{4}{c}{\emph{PACS} (1.5K Samples/Domain)}\\
    \hline
    \!\!Photo-Only\!\! & 79.37\tpm0.17 & 76.16\tpm0.34 & 82.57\tpm0.00 \\
    \!\!Photo + SACL\!\! & 82.35\tpm0.37 & 87.27\tpm0.10 & 77.43\tpm0.84 \\
    \!\!Photo + Painting\!\! & 82.54\tpm0.59 & 79.65\tpm0.49 & \textbf{85.43}\tpm0.70 \\
    \!\!Photo + SACL + Painting\!\! & \textbf{85.42}\tpm0.18 & \textbf{87.31}\tpm0.30 &  83.52\tpm0.27 \\
    \hline

  \end{tabular} 

  \caption{ \textbf{Learning from Stylization and Paintings.} Training with both
  stylized images and paintings improves average robustness to image corruptions
  and out-of-distribution photos, indicating that the invariances learned from
  these images are complementary. \tpm~indicates standard deviation over 3
  runs.} \vspace{-2mm}

  \label{table:comp}
\end{table}

\subsection{The Role of High Frequency Signals}

We have focused our intuitions about the source of invariances learned from
stylization and paintings through the visible structure of these images.
Existing work has shown that CNNs can learn to extract features from high
frequency signals in images \cite{wang2020high, adv_noise}. It is also
well-known that deconvolutional decoders, such as those used in stylization
models, can introduce artifacts in images \cite{checkerboard}. It is difficult
to form intuitions about these signals, but we can measure whether they play a
significant role in improving model robustness.

We apply an ideal circular low-pass filter to zero out
high-frequency components.  Given an image $I$, the filtered frequency
components of the image are:

\vspace{-2mm}
\begin{align}
  X_{\text{filtered}} &= \mathcal{F}(I) \odot C \\
  \nonumber \text{where } C_{ij} &= \mathbf{1}_{r < \tau}({r(i,j)}) 
\end{align}

\begin{figure}[!ht] \centering
  \includegraphics[width=0.8\linewidth]{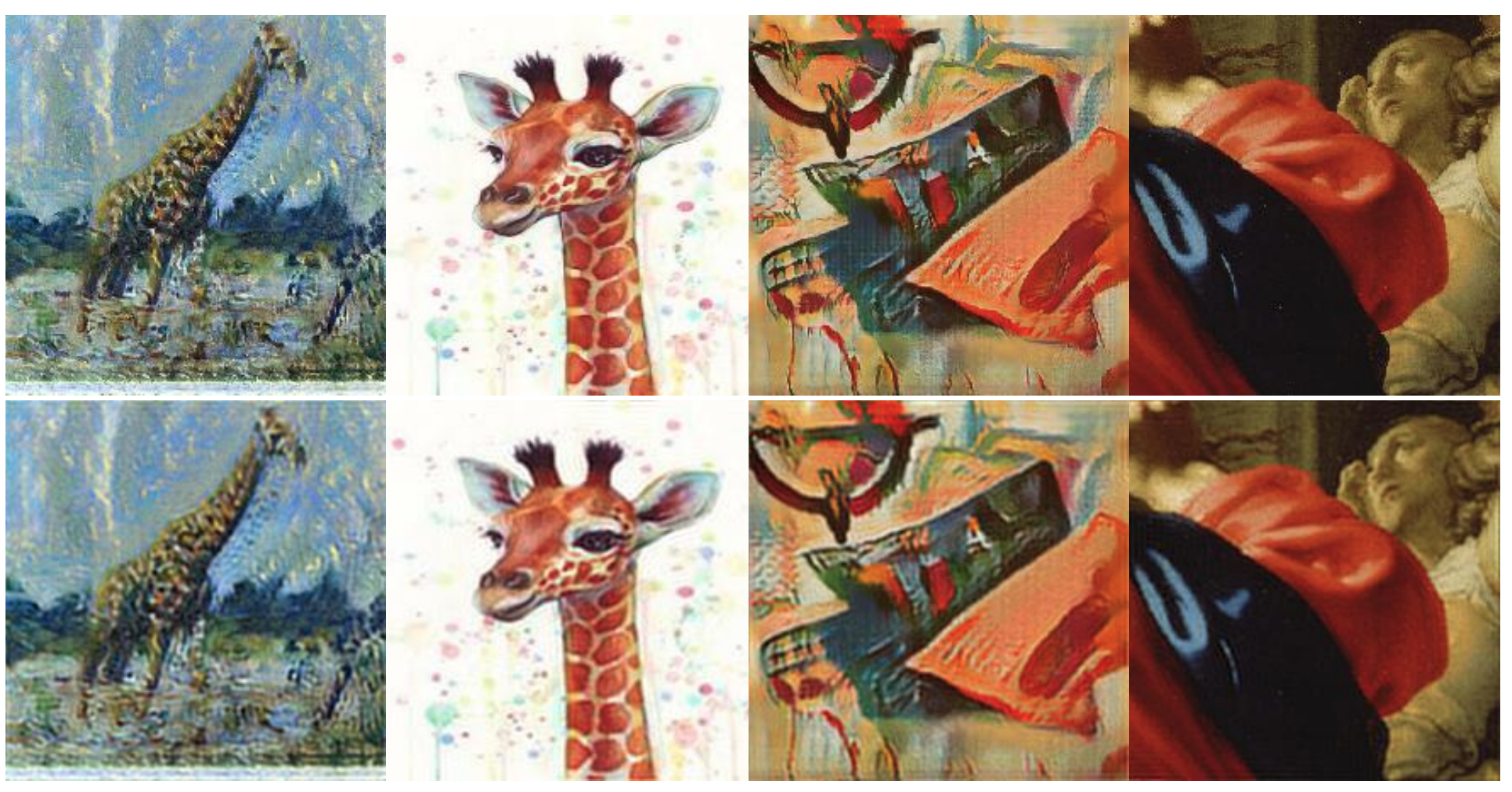}
  \caption{\textbf{Reducing High-Frequency Signals.} Top: Original Image,
  Bottom: Low Frequency Image.  Columns 1 and 3 are stylized photos; columns 2
  and 4 are artist-created paintings.  Reducing the magnitude of sufficiently
  high frequency components from images does not alter perceptual quality of
  images. At a glance, the top and bottom images are perceived to be identical.}
  \vspace{-2mm}

    \label{fig:low_freq}
\end{figure}

$\mathcal{F}$ denotes the discrete 2D Fourier transform, $\mathbf{1}$ denotes
the indicator function, and $\tau$ is the radius of the low-pass filter. We set
$\tau=60$ in our experiments. 
Fig. \ref{fig:low_freq} illustrates images before and after filtering at image
resolution $224\times224$. Note that the filtered images are perceptually
identical to the original images at a glance. Therefore, we can train models on
the filtered images to measure the impact of the visually negligible high
frequency signals which were filtered out. 

Table \ref{table:low_freq} summarizes the results. With filtered images,
robustness against noise drops significantly for models trained on photos
stylized with SACL. This means visible high frequency textures (such as the
brush strokes in a Monet stylized photo) are not enough to explain robustness
against noise. This effect of invisible high-frequency signals on noise is
similar to evidence that learning from adversarial perturbations improves
robustness to high frequency corruptions \cite{yin2019fourier}.  On the other
hand, the effect of high frequency signals on the noise robustness of paintings
is much smaller.

\noindent \textbf{Answer to H6:} \emph{For learned style transfer, it
is the presence of invisible high frequency signals that are doing the heavy
lifting against noise. In contrast, paintings are primarily improving invariance
towards noise through visible human-perceivable signals}. 



%
%



\begin{table}[h!]
  \footnotesize
  \centering
  \begin{tabular}{l||c|c|c|c||c}
    \hline
      \!\!Method\!\! & \makecell[c]{Noise} & \makecell[c]{Blur} &
      \makecell[c]{Weather} &\makecell[c]{Digital} & \makecell[c]{OOD}    \\
    \hline
    \multicolumn{6}{c}{\emph{Materials} (30K Samples/Domain)}\\
    \hline
    \!\!Photo-Only\!\! & 43.70 & 58.76 & 55.25 & 61.20 & 41.33\\
    \hline
    \!\!Photo+SACL\!\! & \textcolor{blue}{61.87} & 64.36 & 57.49 & 66.55 & 34.54 \\
    \!\!Photo+Painting\!\! & \textcolor{red}{49.82} & 61.03 & 56.68 & 64.15 & 43.92 \\
    \hline
    \!\!Photo+SACL (LF)\!\! & \textcolor{blue}{45.82} & 64.24 & 57.06 & 66.37 & 36.92 \\
    \!\!Photo+Painting (LF)\!\! & \textcolor{red}{44.95} & 60.87 & 56.82 & 63.69 & 41.21 \\
    \hline
    \multicolumn{6}{c}{\emph{PACS} (1.5K Samples/Domain)}\\
    \hline
    \!\!Photo-Only\!\! & 62.64 & 72.75 & 83.24 & 86.33 & 82.57\\
    \hline
    \!\!Photo+SACL\!\! & \textcolor{blue}{85.98} & 84.61 & 89.73 & 88.74 & 77.43 \\
    \!\!Photo+Painting\!\! & \textcolor{red}{68.04} & 74.72 & 86.26 & 86.92 & 85.43\\
    \hline
    \!\!Photo+SACL (LF)\!\! & \textcolor{blue}{77.55} & 85.4 & 88.93 & 88.53 & 77.43 \\
    \!\!Photo+Painting (LF)\!\! & \textcolor{red}{71.16} & 75.97 & 86.82 & 87.35 & 83.71\\
    \hline

  \end{tabular} 

  \caption{ \textbf{Robustness without High Frequency Signals.} ``LF'' denotes filtered low frequency images. Photos are always
  unfiltered. Filtering invisible high frequency components mainly impacts noise
  robustness. (\textcolor{blue}{blue}) Filtering stylized
  photos significantly reduces noise robustness while (\textcolor{red}{red})
  filtering paintings has a relatively smaller effect. See supplementary for
  standard deviations.}

  \label{table:low_freq}
\end{table}

%

\section{Conclusion}

In this paper, we have performed an extensive exploration of style transfer and
artistic paintings for model robustness. We found that style transfer is able to
improve model robustness \emph{without} painting style images at all
(\textbf{H1}).  Instead, stylization relies on a combination of diversity
between style-content image pairs and learned biases to improve model robustness
(\textbf{H2}). We further proposed the direct use of paintings as a form of
perceptual data augmentation.  This property of paintings is not easily found
from artforms such as sketches or cartoons due to the fine balance of
abstraction and realism in paintings (\textbf{H4}). We showed that learning from
real paintings can improve robustness, with greater gains found by accounting
for the domain shift between paintings and photos (\textbf{H3}).  Finally, we
found that models learn different invariances from paintings and stylized
photos, and that robustness can be improved by training on both forms of data
(\textbf{H5},\textbf{H6}).

From a practical standpoint, our results suggest that learned stylization
methods should be considered over arbitrary style transfer methods in data
augmentation pipelines. Our results also suggest that training with paintings is
a straightforward way to improve model robustness, and should be used if they
are available.

There are interesting research directions for future exploration. Work has been
done to improve the controls available in style transfer or image editing models
\cite{collins2020editing, wang2018high, gatys2017controlling}. It would be
interesting to apply these controls in a perceptually-grounded manner when style
transfer is applied to mimic the artistic process. In this paper, we have found
that artforms like sketches are unable to improve model robustness. It would be
interesting to explore how coarser abstractions found in art can be
leveraged for model robustness, perhaps by encouraging models to learn a
hierarchy of invariances.  


%

\mypara{Acknowledgements.}
Thanks to the anonymous reviewers for their constructive feedback. This work was
funded in part by NSF (CHS-1617861, CHS-1513967, CHS-1900783, CHS-1930755),
NSERC (PGS-D 516803 2018), and the Netherlands Organization for Scientific
Research (NWO) project 276-54-001.

\clearpage
\renewcommand{\thesection}{A\arabic{section}}
\setcounter{section}{0}

\section{Outline}

 In this supplementary, we provide additional details to enable reproducibility,
 and additional visualizations and results to complement the main findings in
 the paper. In Section \ref{sec:data_appdx}, we specify the creation of the
 Materials dataset. In Section \ref{sec:exp}, we detail the experimental setup
 for the classification robustness experiments. In Section \ref{sec:style_appdx}, we describe the
 implementation and parameters used for style transfer, and we
 show visualizations of these methods in Section \ref{sec:style_vis}. In Section
 \ref{sec:style_dist}, we extend the discussion in Section 4 of the main
 paper by analyzing the effect of stylization strength versus robustness. In
 Section \ref{sec:power}, we visualize the power spectra of stylized images and
 compare them to natural images. Finally, in Section \ref{sec:DA}, we frame
 model robustness as domain generalization and discuss how domain-invariance can
 affect model robustness.

\section{Materials Dataset Details \label{sec:data_appdx}}

In Section 3.2 of the main paper, we briefly described the two primary datasets
on which we focused our experiments. PACS \cite{pacs} is a standard benchmark
dataset while Materials is a novel dataset of photographs and paintings that was
created by sampling image patches from existing datasets with material
annotations. In this section, we give additional information on the creation of
Materials.  This dataset is released for reproducibility at
\small{\url{https://github.com/hubertsgithub/style\_painting\_robustness}}

\paragraph{Natural photographs. }

We acquired image patches from OpenSurfaces\cite{opensurfaces}, COCO stuff
\cite{cocostuff}, and MINC-2500 \cite{minc}. To create image patches for image
classification from segmentation annotations, we constructed bounding boxes
around segments, and cropped out these bounding boxes to form image patches.  We
constructed square bounding boxes with side length equal to 150\% of the minimum
side length of tight bounding box around the segment.  Non-tight bounding boxes
are used since it is important to include some context for the patch. We also
sampled from MINC-2500 which already contains annotated image patches that do
not require additional processing.  Image crops that extend beyond the boundary
of the full image are padded to square with ImageNet mean padding, and all final
images patches are resized to 224$\times$224.  We sampled from OpenSurfaces and
MINC first, before sampling from COCO if necessary.  We created subsets of data
of up to 60K photos, and each subset was created to be
as-class-balanced-as-possible. For illustration, we provide per-class counts for
two such subsets of data in Table \ref{table:dataset}.

\begin{table}[h!]
    \small

    \centering
    \begin{tabular}{ll || ll}

    \hline
    Natural-10K         & Count     & Natural-60K       & Count     \\
    \hline
    Ceramic & 1000 & Ceramic & 3132** \\
    Fabric & 1000 & Fabric & 8006 \\
    Foliage & 1000 & Foliage & 8006 \\
    Glass & 1000 & Glass & 7216** \\
    Liquid & 1000 &  Liquid & 7174** \\
    Metal & 1000 &  Metal & 7204** \\
    Paper & 1000 &  Paper & 3258** \\
    Skin & 1000 &  Skin & 2276** \\
    Stone & 1000 &  Stone & 5716** \\
    Wood & 1000 &  Wood & 8006 \\
    \hline

    \end{tabular}
  \caption{Training datasets are sampled to be as class-balanced as possible. ** indicates that all training samples of that category are included in the training set, and no further samples exist. Natural-10K is a subset of Natural-60K. The test set contains 200 samples of each category.}
    \label{table:dataset}
\end{table}

\mypara{Paintings. } We sample paintings across the same material categories as
above from \cite{van2020database}.  We only sample patches that are at least
128$\times$128 pixels in area to avoid very low-resolution annotations.  The
image patches are padded and resized in the same manner as above, and data is
also sampled to be as-class-balanced-as-possible.

\section{Classification Parameters \label{sec:exp}}

For all classification experiments, we use the following setup. Code is
released for reproducibility at 
\small{\url{https://github.com/hubertsgithub/style\_painting\_robustness}}

\begin{itemize}
    \vspace{-2mm}
  \item Network architecture: ResNet18, ImageNet pretrained.
    \vspace{-2mm}
  \item Training hyperparameters: 30 epochs with initial learning rate (LR)
    1e-3, LR reduced to 1e-4 at epoch 24. The LR of the classification layer is
    increased by 10$\times$.
    \vspace{-2mm}
  \item Optimizer: SGD with 0.9 momentum. 
    \vspace{-2mm}
  \item Training data augmentation: horizontal flipping, random scaling, color jitter, and ImageNet normalization.
  \vspace{-2mm}
  \item For experiments that train a model on both photos and stylized photos,
    all photos are stylized exactly once offline and
    included in the training set as an independent image from the original photo.
  \vspace{-2mm}
  \item Evaluation accuracies are averaged over 3 independent runs for each
    experiment. 
    \vspace{-2mm}
\end{itemize}

Our results do not appear sensitive to the choice of training
hyperparameters. Therefore, we train all networks with this configuration and
evaluate the final model after training.  Our experiments suggest that this
training schedule is sufficient for convergence without overfitting across all
datasets we experimented with. Increasing training epochs to 100 or more does
not improve results. Increasing number of training epochs is required if
starting from random initialization, but ImageNet pretraining is standard
practice so we do not extensively experiment with random initialization.

\section{Style Transfer Parameters \label{sec:style_appdx}}

For all applications of style transfer used in this work, we use pretrained
models from publicly available implementations. The sources are provided here:

\begin{itemize}
  \item AdaIN \cite{adain}: \texttt{https://github.com/bethgelab/\\stylize-datasets}
  \item ETNet \cite{etnet}: \texttt{https://github.com/zhijieW94/\\ETNet}
  \item TPFR \cite{tpfr}: \texttt{https://github.com/nnaisense/\\conditional-style-transfer}
  \item SACL \cite{sacl}: \texttt{https://github.com/CompVis/\\adaptive-style-transfer}
\end{itemize}

Our initial experiments showed that applying style transfer at 224$\times$224
resolution yielded visually poor results (except for AdaIN). Therefore, we apply
style transfer at a higher resolution and downsample the final result to
224$\times$224. For AdaIN, ETNet, and SACL, we apply style transfer at
768$\times$768 resolution. For TPFR, we apply style transfer at 512$\times$512
instead of 768$\times$768 due to GPU memory constraints.  All other
hyperparameters are set to the default settings found in the implementations for
each respective method.

\section{Visualizations of Stylized Photos \label{sec:style_vis}}

We show examples of images stylized by various style transfer methods on PACS
(Fig.  \ref{fig:style_vis_PACS}) and Materials (Fig.
\ref{fig:style_vis_mat}). The visualizations also include examples of
intradomain stylization in which images are stylized by photos instead of by
paintings.  Notice that intradomain stylization yields stylizations that are, in
general, visually similar to stylizations with painting style images. Overall,
stylizations across all methods are holistically similar to natural paintings.

\section{Style Distance vs Robustness \label{sec:style_dist}}

In Section 4.2, we found that arbitrary stylization with style images that share
the same semantic content as the content image (``intraclass stylization'')
results in lower gains in robustness. Since images with similar semantic content
may be more visually similar, this suggests that intraclass stylization will
lead to less stylized images, i.e. weaker augmentation.  To verify this, we
measured style differences via the Gram matrix distance between stylized images
and their original counterparts. Table \ref{table:style_appdx} summarizes differences
on PACS. While intraclass stylization does result in smaller differences in
style for each method, the Gram matrix distance across methods is not
necessarily correlated with gains in robustness. For example, ETNet produces the
largest style differences overall, but AdaIN improves robustness more (Fig.
3,4). As such, the strength of stylization alone is not indicative of the
downstream robustness learned by models trained on these images.

\begin{table}[h!]
  \small
  \centering
  \begin{tabular}{l||c|c|c}
    \hline
      \!\!Method\!\! &
      \makecell[c]{Painting}& \makecell[c]{Intradomain} & \makecell[c]{Intradomain \\
      (Intraclass)}   \\
    \hline
    \!\!AdaIN\!\! & 1.58\tpm0.93 &  1.28\tpm0.79 & 1.16\tpm0.85 \\
    \!\!ETNet\!\! & 2.33\tpm1.09 & 2.13\tpm1.04 & 1.81\tpm1.03  \\
    \!\!TPFR\!\! &  1.52\tpm0.90 & 1.38\tpm0.87 & 1.27\tpm0.91  \\
    \hline
  \end{tabular} 

  \caption{\textbf{Style (Gram Matrix) Distance.} Gram matrices
  computed from ImageNet pretrained ResNet18 features on PACS. Mean distance between
  \texttt{(image, stylized image)} pairs is reported. $\uparrow$ distance
  implies $\uparrow$ style difference. \tpm~denotes standard deviation
  across 1.5K pairs.}
  
  \label{table:style_appdx}
\end{table}

\section{Power Spectra of Different Image Types \label{sec:power}}

In Section 6 of the main paper, we found that SACL improves robustness against
noise with imperceptible high frequency signals in the stylized images. The
results are shown in Table \ref{table:low_freq_full}. Here we
show the power spectra of stylized images and compare them to the spectra for
natural photos and natural paintings. The radial power spectrum for an image is
computed as:

\begin{align}
  \nonumber \text{power}(r) &= ||X_{r}||^2  \\
  \nonumber \text{where } X_{r} &= \mathop\text{mean}\limits_{\sqrt{i^2 + j^2} \in
  \mathcal{R}(r)}||X_{ij}||
\end{align}

\begin{figure}[!ht] \centering
  \includegraphics[width=\linewidth]{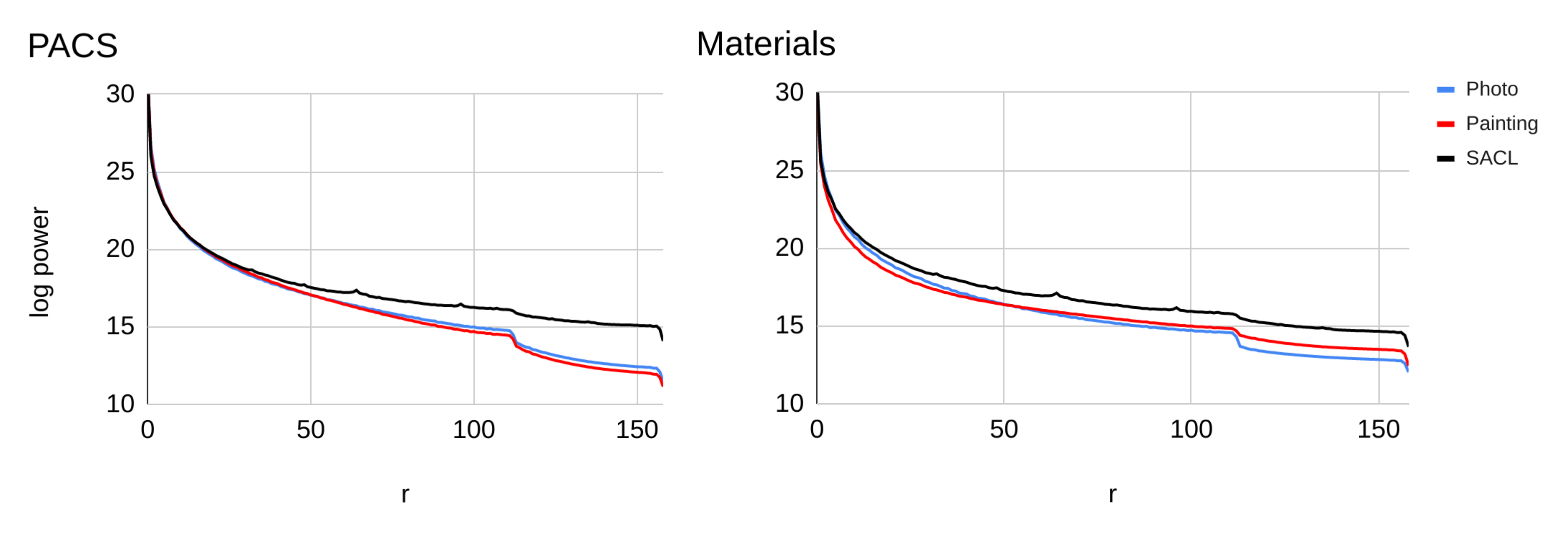} \caption{\textbf{Power
  Spectrum of Images}. Left: PACS, Right: Materials. The plots depict the mean
  power spectrum for different sets of images. Photos stylized by SACL have
  larger magnitude high frequency components than natural photos or natural
  paintings. }

    \label{fig:radial_power}
\end{figure}

\noindent $X_{ij}$ are the frequency components given by the 2D discrete Fourier
transform. Since $(i,j)$ are discrete, the radial frequency component $X_r$ is
computed as an average over $||X_{ij}||$ for ${(i,j)}$ that fall in a bin
$\mathcal{R}(r)$. In Fig.  \ref{fig:radial_power}, we visualize the mean radial
power spectra for natural photos, natural paintings, and SACL-stylized photos.
We observe that stylized photos contain higher magnitude high-frequency
components relative to natural photos and natural paintings. As noted in Section
6 of the main paper, reducing the magnitude of sufficiently high-frequency
components does not affect the perceptual quality of images.

\begin{table*}[h!]
  \small
  \centering
  \begin{tabular}{l||c|c|c|c||c}
    \hline
      \!\!Method\!\! & \makecell[c]{Noise} & \makecell[c]{Blur} &
      \makecell[c]{Weather} &\makecell[c]{Digital} & \makecell[c]{OOD}    \\
    \hline
    \multicolumn{6}{c}{\emph{Materials} (30K Samples/Domain)}\\
    \hline
    \!\!Photo-Only\!\! & 43.70\tpm0.65 & 58.76\tpm0.14 & 55.25\tpm0.33 &
    61.20\tpm0.69 & 41.33\tpm0.62  \\
    \hline
    \!\!Photo + SACL\!\! & \textcolor{blue}{ 61.87}\tpm0.16 & 64.36\tpm0.20 &
    57.49\tpm0.24 & 66.55\tpm0.17 & 34.54\tpm0.91\\
    \!\!Photo + Painting\!\! & \textcolor{red}{49.82}\tpm0.56 & 61.03\tpm0.13 &
    56.69\tpm0.10 & 64.15\tpm0.14 & 43.92\tpm0.47\\
    \hline
    \!\!Photo+SACL (LF)\!\! & \textcolor{blue}{45.82}\tpm1.36 & 64.24\tpm0.39 &
    57.06\tpm0.13 & 66.37\tpm0.29 & 36.92\tpm1.15 \\
    \!\!Photo+Painting (LF)\!\! & \textcolor{red}{44.95}\tpm0.66 & 60.87\tpm0.29
    & 56.82\tpm0.23 & 63.69\tpm0.46 & 41.21\tpm0.56 \\
    \hline
    \multicolumn{6}{c}{\emph{PACS} (1.5K Samples/Domain)}\\
    \hline
    \!\!Photo-Only\!\! & 62.64\tpm1.48 & 72.75\tpm0.04 & 83.24\tpm0.22 &
    86.33\tpm0.14 & 82.57\tpm0.00\\
    \hline
    \!\!Photo + SACL\!\! & \textcolor{blue}{85.98}\tpm0.56 & 84.61\tpm0.15 &
    89.73\tpm0.33 & 88.74\tpm0.48 & 77.43\tpm0.84\\
    \!\!Photo + Painting\!\! & \textcolor{red}{68.83}\tpm0.83 & 75.80\tpm0.95 &
    86.88\tpm0.66 & 87.07\tpm0.14 & 85.43\tpm0.70 \\
    \hline
    \!\!Photo+SACL (LF)\!\! & \textcolor{blue}{77.55}\tpm2.60 & 85.4\tpm0.11 &
    88.93\tpm0.22 & 88.53\tpm0.15 & 77.43\tpm0.47 \\
    \!\!Photo+Painting (LF)\!\! & \textcolor{red}{71.16}\tpm1.31 & 75.97\tpm0.71
    & 86.82\tpm0.37 & 87.35\tpm0.36 & 83.71\tpm0.40\\
    \hline

  \end{tabular} 

  \caption{ \textbf{Robustness without High Frequency Signals.} ``LF'' denotes filtered low frequency images. Photos are always
  unfiltered. Filtering invisible high frequency components mainly impacts noise
  robustness. (\textcolor{blue}{blue}) Filtering stylized
  photos significantly reduces noise robustness while (\textcolor{red}{red})
  filtering paintings has a relatively smaller effect. \tpm~indicates standard
  deviations over 3 runs.}

  \label{table:low_freq_full}
\end{table*}



\section{Domain-Invariant Feature Learning \label{sec:DA}}

Our results from Section 5 of the main paper provide evidence that models can
learn more robust feature representations from the addition of paintings to a
dataset of photographs. We can take this further by explicitly enforcing similar
(or domain-invariant) feature representations across photos and paintings.
Domain-invariance is a common approach to the problem of domain generalization,
where models are trained on multiple domains with the goal of generalizing to
unseen domains, e.g. \cite{gulrajani2020search,mmld}. In our setting, we can consider
images with common corruptions to be the set of unseen domains.  Perfect
domain-invariant feature extraction can be harmful if it prevents useful
features in photos from being extracted due to an underrepresentation of such
features in paintings. Since the target task is recognition of photos, losing
robust photo-specific signals can be detrimental. Therefore, we explore the
following: 

\begin{itemize}
\item \textbf{Hypothesis H1S:} Explicitly learning
domain-invariance from paintings
  and photos may negatively impact model robustness.
\end{itemize}

We use an adversarial domain discriminator to learn domain invariant features
\cite{ganin, mmld}. In Table \ref{table:DA}, we find that explicitly learning
domain invariant features from paintings results in lower robustness than
unrestricted feature learning with paintings. However, learning domain-invariant
features does still improve robustness over the photo-only baseline. Existing
work in domain generalization has shown that domain-invariance is an effective
method for learning to recognize images from unseen domains, e.g. \cite{mmld}.
Our finding here suggests that in the special case of domain generalization to
corrupted versions of natural photographs, it is advantageous to retain
photo-specific features for recognition. This is consistent with our hypothesis
and discussion above -- an underrepresentation of any particular  photo-specific
features in paintings can result in such features being ignored entirely when
domain-invariance is enforced, even if such features are useful for robust
recognition.

\noindent \textbf{Answer to H1S:} \emph{Explicitly learning domain-invariant
features from paintings negatively impacts model robustness with respect to
unrestricted feature learning with paintings. However, domain-invariant features
do still improve robustness relative to photos only.}


\begin{table}[h!]
  \footnotesize
  \centering
  \begin{tabular}{l||c||c|c|c|c}
    \hline
      \!\!Method\!\! & \makecell[c]{\!\!MEAN\!\!} & \makecell[c]{\!\!Noise\!\!}
      & \makecell[c]{\!\!Blur\!\!} &
      \makecell[c]{\!\!Weather\!\!} &\makecell[c]{\!\!Digital\!\!}    \\
    \hline
    \multicolumn{5}{c}{\emph{Materials} (30K Samples/Domain)}\\
    \hline
    \!\!Photo-Only\!\! & 54.73  & 43.71 & 58.76 & 55.25 & 61.20 \\
    \makecell[l]{\!\!Photo + Painting\!\!} & \textcolor{blue}{\bf 57.92} &
    \textbf{49.82} & \textbf{61.03} & \textbf{56.68} & \textbf{64.15} \\
    \makecell[l]{\!\!Photo + Painting (DA)\!\!} & \textcolor{red}{55.99} &  46.97 & 59.60 & 54.51 & 62.90 \\
    \hline
    \multicolumn{5}{c}{\emph{PACS} (1.5K Samples/Domain)}\\
    \hline
    \!\!Photo-Only\!\! & 76.16 & 62.64 & 72.75 & 83.24 & 86.33 \\
    \!\!Photo + Painting\!\! & \textcolor{blue}{\bf 78.99} &  68.04 &
    \textbf{74.72} & \textbf{86.26} & \textbf{86.92} \\
    \makecell[l]{\!\!Photo + Painting (DA)\!\!} & \textcolor{red}{77.44}  &
    \textbf{68.86} & 72.59 & 84.09 & 84.23 \\
    \hline

  \end{tabular} 

  \caption{ \textbf{Effect of Domain-Invariant Features.} ``DA'' refers to
  feature learning with an adversarial domain discriminator loss \cite{ganin}.
  Learning domain-invariant features (\textcolor{red}{red}) reduces robustness
  relative to unrestricted feature learning from paintings
  (\textcolor{blue}{blue}), but still improves robustness over photo-only.}

  \label{table:DA}
\end{table}

\section{Additional Architectures}

 We expect our findings to hold across architectures and datasets. As a sanity
 check, we have extended Table 2 with two additional architectures. The results
 (Table \ref{table:table2_extended}) follow similar trends to those found in
 Table 2.  For example, SACL  outperforms both AdaIN and
 Paintings on Noise.

\begin{table*}[h!] 
  \small
  \centering

  \textbf{Resnet-18}\\
  \begin{tabular}{l||c|c|c|c}
    \hline
      \!\!Method\!\! & \makecell[c]{Noise} & \makecell[c]{Blur} &
      \makecell[c]{Weather} &\makecell[c]{Digital}    \\
    \hline
    \multicolumn{5}{c}{\emph{Materials} (30K Samples/Domain)}\\
    \hline
    \!\!Photo-Only\!\! & 43.70\tpm0.65 & 58.76\tpm0.14 & 55.25\tpm0.33 &
    61.20\tpm0.69 \\
    \hdashline
    \!\!Photo + AdaIN\!\! & \textcolor{red}{47.33}\tpm0.22 & 65.09\tpm0.21 &
    61.78\tpm0.18 & 61.41\tpm0.16\\
    \!\!Photo + SACL\!\! & \textcolor{blue}{61.87}\tpm0.16 & 64.36\tpm0.20 &
    57.49\tpm0.24 & 66.55\tpm0.17 \\
    \!\!Photo + Painting\!\! & \textcolor{red}{49.82}\tpm0.56 & 61.03\tpm0.13 &
    56.69\tpm0.10 & 64.15\tpm0.14 \\
    \hline
    \multicolumn{5}{c}{\emph{PACS} (1.5K Samples/Domain)}\\
    \hline
    \!\!Photo-Only\!\! & 62.64\tpm1.48 & 72.75\tpm0.04 & 83.24\tpm0.22 &
    86.33\tpm0.14 \\
    \hdashline
    \!\!Photo + AdaIN\!\! & \textcolor{red}{70.17}\tpm1.70 & 81.18\tpm0.20 &
    88.37\tpm0.23 & 89.32\tpm0.19 \\
    \!\!Photo + SACL\!\! & \textcolor{blue}{85.98}\tpm0.56 & 84.61\tpm0.15 &
    89.73\tpm0.33 & 88.74\tpm0.48 \\
    \!\!Photo + Painting\!\! & \textcolor{red}{68.83}\tpm0.83 & 75.80\tpm0.95 &
    86.88\tpm0.66 & 87.07\tpm0.14 \\
    \hline
  \end{tabular} 

  \vspace{2mm}
  \textbf{WideResnet-50-2}\\
  \begin{tabular}{l||c|c|c|c}
    \hline
      \!\!Method\!\! & \makecell[c]{Noise} & \makecell[c]{Blur} &
      \makecell[c]{Weather} &\makecell[c]{Digital}    \\
    \hline
    \multicolumn{5}{c}{\emph{Materials} (30K Samples/Domain)}\\
    \hline
    \!\!Photo + AdaIN\!\! & \textcolor{red}{57.80}\tpm1.79 &	73.77\tpm0.11 &	67.75\tpm0.51 &
    66.96\tpm0.06 \\
    \!\!Photo + SACL\!\! & \textcolor{blue}{69.39}\tpm0.72 & 70.00\tpm0.34 & 64.00\tpm0.54 &
    73.05\tpm0.30 \\
    \!\!Photo + Painting\!\! & \textcolor{red}{60.72}\tpm0.83 & 68.09\tpm0.49 & 61.15\tpm0.23
    & 70.98\tpm0.24 \\
    \hline
    \multicolumn{5}{c}{\emph{PACS} (1.5K Samples/Domain)}\\
    \hline
    \!\!Photo + AdaIN\!\! & \textcolor{red}{82.05}\tpm1.33 &	86.89\tpm0.64 &	93.98\tpm0.15 &
    94.39\tpm0.30 \\
    \!\!Photo + SACL\!\! &  \textcolor{blue}{93.79}\tpm1.35 & 89.64\tpm0.36 & 95.19\tpm0.17 &
    93.63\tpm0.11 \\
    \!\!Photo + Painting\!\! & \textcolor{red}{83.92}\tpm1.81 & 85.38\tpm0.27 & 94.19\tpm0.08
    & 92.63\tpm0.24 \\
    \hline
  \end{tabular} 

  \vspace{2mm}
  \textbf{Densenet-121}\\
  \begin{tabular}{l||c|c|c|c}
    \hline
      \!\!Method\!\! & \makecell[c]{Noise} & \makecell[c]{Blur} &
      \makecell[c]{Weather} &\makecell[c]{Digital}    \\
    \hline
    \multicolumn{5}{c}{\emph{Materials} (30K Samples/Domain)}\\
    \hline
    \!\!Photo + AdaIN\!\! & \textcolor{red}{54.32}\tpm0.23 &	71.08\tpm0.24 &	67.31\tpm0.37 &
    66.47\tpm0.13 \\
    \!\!Photo + SACL\!\! &  \textcolor{blue}{67.22}\tpm0.16 & 68.89\tpm0.16 & 63.08\tpm0.33 &
    71.87\tpm0.62 \\
    \!\!Photo + Painting\!\! & \textcolor{red}{54.83}\tpm1.20 & 68.21\tpm0.38 & 61.29\tpm0.39
    & 70.66\tpm0.13 \\
    \hline
    \multicolumn{5}{c}{\emph{PACS} (1.5K Samples/Domain)}\\
    \hline
    \!\!Photo + AdaIN\!\! & \textcolor{red}{76.96}\tpm4.12 &	85.79\tpm0.50 &	94.96\tpm0.13 &
    92.34\tpm0.19 \\
    \!\!Photo + SACL\!\! &  \textcolor{blue}{91.33}\tpm0.28 & 88.92\tpm0.37 & 94.18\tpm0.49 &
    94.12\tpm0.54 \\
    \!\!Photo + Painting\!\! & \textcolor{red}{76.65}\tpm2.22 & 83.22\tpm0.19 & 94.00\tpm0.62
    & 91.72\tpm0.14 \\
    \hline
  \end{tabular} 

  \caption{ \textbf{Per-Corruption Accuracy (Additional Architectures).} Trends
  across different architectures are generally consistent. For example, SACL (\textcolor{blue}{blue}) greatly outperforms
  AdaIN and paintings (\textcolor{red}{red}) for noise robustness. \tpm~indicates standard
  deviation over 3 runs.} 

  \label{table:table2_extended}
\end{table*}

\begin{figure*}[!ht] 
  \centering 
  \includegraphics[width=\linewidth]{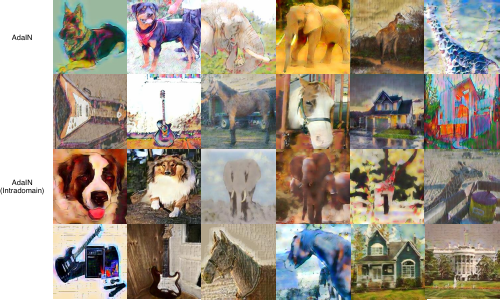}
  \includegraphics[width=\linewidth]{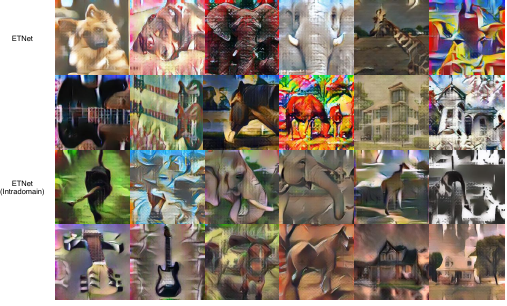}

  \caption{\textbf{Stylized Photos (PACS) (1/2)}. Intradomain refers to stylization with
  photos as style images instead of paintings as style images. SACL is a learned
  style transfer method that is applied with different models pretrained to
  transfer the style of different artists. (\emph{Continued on next page})}

    \label{fig:style_vis_PACS}
\end{figure*}

\begin{figure*}[!ht] 
  \ContinuedFloat
  \centering 
  \includegraphics[width=\linewidth]{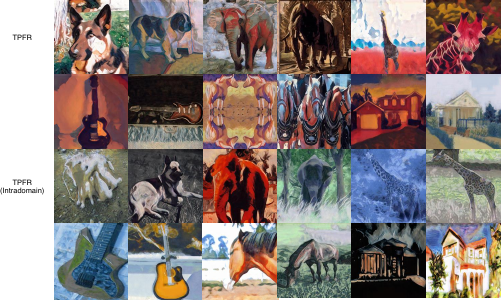}
  \hspace*{3.5mm}\includegraphics[width=0.985\linewidth]{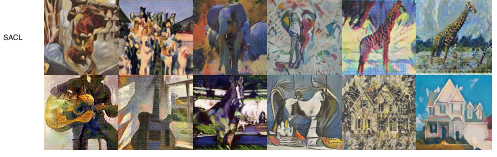}

  \caption{\textbf{Stylized Photos (PACS) (2/2)}. Intradomain refers to
  stylization with photos as style images instead of paintings as style images.
  SACL is a learned style transfer method that is applied with different models
  pretrained to transfer the style of different artists.}

\end{figure*}

\begin{figure*}[!ht] 
  \centering 
  \includegraphics[width=\linewidth]{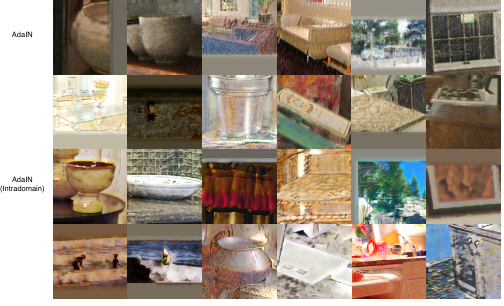}
  \includegraphics[width=\linewidth]{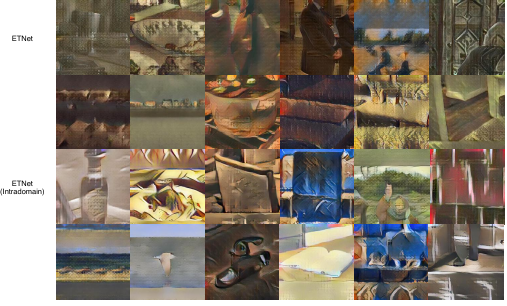}

  \caption{\textbf{Stylized Photos (Materials) (1/2)}. Intradomain refers to stylization with
  photos as style images instead of paintings as style images. SACL is a learned
  style transfer method that is applied with different models pretrained to
  transfer the style of different artists. (\emph{Continued on next page})}

    \label{fig:style_vis_mat}
\end{figure*}

\begin{figure*}[!ht] 
  \ContinuedFloat
  \centering 
  \includegraphics[width=\linewidth]{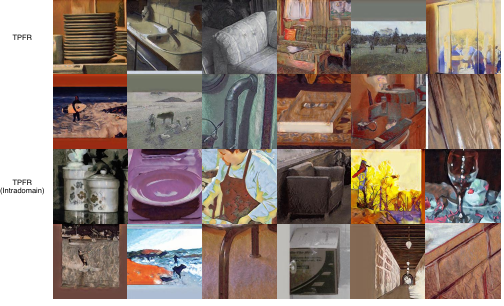}
  \hspace*{3.5mm}\includegraphics[width=0.985\linewidth]{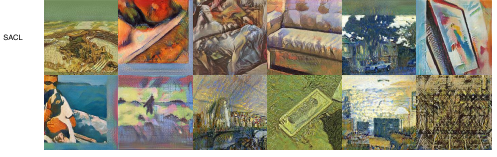}

  \caption{\textbf{Stylized Photos (Materials) (2/2)}. Intradomain refers to
  stylization with photos as style images instead of paintings as style images.
  SACL is a learned style transfer method that is applied with different models
  pretrained to transfer the style of different artists.}

\end{figure*}

\clearpage
\newpage
{\small
\bibliographystyle{ieee_fullname}
\bibliography{paper}
}

\newpage


\end{document}